\documentclass[conference]{IEEEtran}
\usepackage{comment}
\IEEEoverridecommandlockouts
\usepackage{cite}
\usepackage{amsmath,amssymb,amsfonts}
\usepackage{algorithmic}
\usepackage{graphicx}
\usepackage{textcomp}
\usepackage{xcolor}
\usepackage{booktabs}
\usepackage{multirow} 
\usepackage{adjustbox}
\usepackage{hyperref}
\usepackage{enumitem}
\usepackage{tabularx}
\usepackage{algorithm}

\def\BibTeX{{\rm B\kern-.05em{\sc i\kern-.025em b}\kern-.08em
    T\kern-.1667em\lower.7ex\hbox{E}\kern-.125emX}}
\begin{document}

\title{Action Space Reduction Strategies for Reinforcement Learning in Autonomous Driving\\

{\footnotesize %\textsuperscript
}
\thanks{This work was supported in part by the National Science Foundation (NSF) under Grant MRI 2214830.}
}

\author{
  \IEEEauthorblockN{ 
  Elahe Delavari}
  \IEEEauthorblockA{
    Electrical and Computer Engineering\\
    University of Michigan–Dearborn\\
    Dearborn, MI, USA\\
    elahed@umich.edu
  }
\and
  \IEEEauthorblockN{ 
  Feeza Khan Khanzada}
  \IEEEauthorblockA{
    Electrical and Computer Engineering\\
    University of Michigan–Dearborn\\
    Dearborn, MI, USA\\
    feezakk@umich.edu
  }
\and
  \IEEEauthorblockN{ 
  Jaerock Kwon}
  \IEEEauthorblockA{
    Electrical and Computer Engineering\\
    University of Michigan–Dearborn\\
    Dearborn, MI, USA\\
    jrkwon@umich.edu
  }
}

\maketitle

\begin{abstract}
Reinforcement Learning (RL) offers a promising framework for autonomous driving by enabling agents to learn control policies through interaction with environments. However, large and high-dimensional action spaces—often used to support fine-grained control—can impede training efficiency and increase exploration costs.  In this study, we introduce and evaluate two novel structured action space modification strategies for RL in autonomous driving: dynamic masking and relative action space reduction. These approaches are systematically compared against fixed reduction schemes and full action space baselines to assess their impact on policy learning and performance. Our framework leverages a multimodal Proximal Policy Optimization agent that processes both semantic image sequences and scalar vehicle states. The proposed dynamic and relative strategies incorporate real-time action masking based on context and state transitions, preserving action consistency while eliminating invalid or suboptimal choices. Through comprehensive experiments across diverse driving routes, we show that action space reduction significantly improves training stability and policy performance. The dynamic and relative schemes, in particular, achieve a favorable balance between learning speed, control precision, and generalization. These findings highlight the importance of context-aware action space design for scalable and reliable RL in autonomous driving tasks.

\end{abstract}


\section{Introduction}

The development of Autonomous Vehicles (AVs) has accelerated in recent years, offering the potential to improve road safety, reduce traffic congestion, and enhance mobility. However, building reliable and efficient self-driving systems remains a formidable challenge due to the complexity of real-world driving. These environments involve dynamic interactions with multiple agents, unpredictable traffic behaviors, and rare but critical edge cases that demand robust decision-making.

Reinforcement Learning (RL) has emerged as a promising paradigm for autonomous driving, enabling agents to learn control policies through interaction with the environment. Unlike rule-based or supervised approaches, RL can adaptively optimize sequential decision-making in tasks such as lane keeping, intersection navigation, and obstacle avoidance~\cite{Learning-to-drive-in-a-day}. However, the application of RL in AVs is often hindered by the large and high-dimensional action spaces inherent to driving tasks. AV control typically requires fine-grained coordination of steering, throttle, and braking actions, represented through either discrete~\cite{Think2Drive,GRI:General_Reinforced_Imitation,End-to-End_Model-Free_Reinforcement_Learning_for_Urban_Driving,elallid_dqn-based_2022,Deep_reinforcement_learning_based_control_for_Autonomous_Vehicles_in_CARLA,Learning_to_drive_from_a_world_on_rails} or continuous~\cite{deng_context-aware_2024,Reinforced_Curriculum_Learning_For_Autonomous_Driving,deng_context_2023,wu_reinforcement_2023,Versatile_and_Efficien,End-to-End_Urban_Driving_by_Imitating_a_Reinforcement_Learning_Coach} action spaces. These large action spaces pose several challenges: they increase computational cost, reduce exploration efficiency, slow policy convergence, and complicate action interpretability. Discrete spaces must balance granularity and tractability~\cite{Tavakoli2017Action}, while continuous spaces often require carefully designed exploration noise and entropy-based regularization to ensure effective learning~\cite{Lillicrap2015Continuous}. Beyond computational issues, large action spaces can lead to erratic or unnatural behaviors, undermining the smoothness and safety required for real-world deployment. Recent studies have highlighted action space reduction as a practical strategy to address these issues~\cite{Learn_What_Not}, though its application to AV domains remains underexplored. By pruning irrelevant or redundant actions, such techniques can accelerate training and promote more stable and human-like policies~\cite{Dynamic_Action_Space,Learn_What_Not}. Approaches such as dynamic elimination~\cite{Dynamic_Action_Space} and state-specific action masking~\cite{Learning_State_Specific} have shown promise in simplifying decision-making without sacrificing performance.

Human drivers intuitively limit their choices based on context—ignoring actions that are unsafe or infeasible in a given moment. In contrast, standard RL agents evaluate all actions at every timestep, often wasting effort on suboptimal options. To address this inefficiency, we propose a structured action space reduction framework that enables RL agents to make context-aware decisions by dynamically filtering available actions. This strategy preserves the flexibility of RL while improving sample efficiency, stability, and policy interpretability.

This work introduces two novel action space modification strategies—dynamic masking and relative reduction—tailored for RL in AV. These are systematically compared against fixed reduction baselines and full action space configurations. In the fixed approach, a predefined subset of steering and throttle actions is used. In contrast, the dynamic and relative strategies adaptively limit the available actions at each timestep based on the agent’s current state. 
Our motivation is rooted in two core challenges: (i) improving training efficiency through structured reduction of the action space to speed up policy convergence, and (ii) enhancing driving performance by fostering smoother, more human-like control. This paper systematically evaluates the impact of these strategies on learning speed, trajectory smoothness, and driving success rates, providing practical insights into designing efficient and robust RL-based AV policies.
\begin{comment}
\begin{figure}[h]
    \centering
    \includegraphics[width=\linewidth, keepaspectratio]{figs/Overview_2.pdf}
    \caption{Overview of the proposed architecture for reinforcement learning-based autonomous driving.}
    \label{fig:overview_architecture}
\end{figure}
\end{comment}

Our proposed framework processes multimodal state representations—including semantic image sequences and scalar vehicle states—through a structured pipeline. A pretrained encoder extracts visual features, which are temporally aggregated using a Conv1D module. Scalar features such as speed, throttle, and steering are encoded separately and fused with the visual representation. The resulting latent state is used by a Proximal Policy Optimization (PPO) agent~\cite{Schulman2017PPO}, with action masking applied using the MaskablePPO variant to enforce valid action constraints during training.

\begin{comment}
To summarize, our main contributions are:
\begin{itemize}
    \item We propose two novel structured action space reduction strategies—\textbf{dynamic} and \textbf{relative}—that use real-time action masking based on the agent’s state to enable adaptive, context-aware control.
    \item We develop a multimodal RL framework that integrates semantic vision and scalar features using a pretrained encoder and temporal feature extractor, trained with PPO and action masking.
    \item We conduct a comprehensive evaluation in the CARLA \cite{CARLA} simulator across diverse driving scenarios. Our results demonstrate that action space reduction improves training speed, lane adherence, and success rate—with dynamic and relative strategies offering the best trade-offs.
\end{itemize}
\end{comment}

To summarize, we tackle the sample-inefficiency that arises with deep-RL agents in AV domain. Building on the intuition that most actions are either kinematically impossible or redundant in a given state, we introduce two novel context-aware action space modification schemes—(i) dynamic masking and (ii) relative reduction—that prune or remap invalid actions on-the-fly while keeping the action dimensionality fixed. We embed these schemes in a multimodal PPO framework that fuses BEV semantic vision with five scalar vehicle states via a MobileNetV3 \cite{howard2019mobilenetv3} encoder and temporal Conv1D head. The relative ±0.5 variant start converging after about 1 M steps—about 2 × faster than the full ±0.5 baseline—while having a comparable success rate to the full action space baseline based on our training and validation in Town07 of CARLA \cite{CARLA} simulator.

%reaches its reward plateau after about 1 M steps—about 2 × faster than the full ±0.5 baseline—while having a comparable success rate to the full action space baseline based on our training and validation in Town07 of CARLA \cite{CARLA} simulator.

%\textcolor{red} {Maybe I need to explain the scenario like lane keeping or whatever for driving. I am not sure if the MDP should be also explained or not.}
%%%%%%%%%%%%%%%%%%%%%%%%%%%

\section{Related Work}
\label{sec:related_work}

In self-driving cars, the action space defines the range of possible decisions an agent can make, including steering, throttle, and braking. It serves as a critical interface between the RL algorithm and the vehicle's control system, directly impacting control precision, learning efficiency, and adaptability in dynamic environments \cite{kiran_deep_2022}. Effective action space design is crucial for enabling smooth maneuvers such as lane changes and obstacle avoidance while ensuring efficient learning in complex driving scenarios \cite{wang_continuous_2019}.

Action spaces in RL can be broadly categorized into discrete, continuous and hybrid designs \cite{li_hyar_2022}. Discrete action spaces simplify learning by constraining decisions to predefined categories, such as fixed combinations of throttle and steering values. Continuous action spaces, in contrast, allow for smooth, precise control by representing actions as values within a range (e.g., [-1,1] for steering) and are widely used for tasks requiring fine-grained adjustments. Hybrid spaces combine discrete and continuous controls, achieving a balance between simplicity and precision.

Despite their importance, large or high-dimensional action spaces pose significant scalability challenges, including slower convergence, increased computational costs, and inefficient exploration \cite{dulac-arnold_deep_2016}. To address these challenges, various action space modification techniques have been proposed that can be categorized into two main categories:
\begin{itemize}
    \item \textit{Action‑space reduction:} Any offline or online transformation that permanently lowers the effective number of actions an agent must consider—e.g.\ pruning infeasible moves, merging similar actions, or re‑encoding them in a lower‑dimensional form—thereby shrinking the branching factor and speeding up learning without (ideally) discarding the optimal solution \cite{li_achieving_2023}.

    \item \textit{Action masking:} A state‑conditioned filter that temporarily disables illegal or irrelevant actions before the policy selects among them, thereby reducing the agent’s effective branching factor without permanently deleting actions from the underlying space \cite{huang_closer_2022}.
\end{itemize}

These action-space reduction and masking techniques have been widely applied both in general RL research and in autonomous-vehicle and mobile-robot navigation. In the following subsections, we review representative studies in each of these areas.

\subsection{Generic RL: Action-Space Reduction and Masking}

\paragraph{Action Reduction}

Kanervisto et al.~\cite{Action_Space_Shaping} proposed action shaping{*}, a static preprocessing step that removes obviously infeasible actions, discretizes continuous dimensions, and merges multi-discrete factors into single discrete choices. In domains such as StarCraft II \cite{vinyals_starcraft_2017} and ViZDoom \cite{kempka_vizdoom_2016}, the approach yields marked gains in computation and sample efficiency, but its static nature prevents it from adapting when the relevance of an action changes over time.  Li et al.~\cite{Achieving_Sample} instead introduce action grouping, partitioning actions according to their transition and reward characteristics. By balancing approximation and estimation errors, the method reduces complexity while preserving optimality in resource-constrained settings.

\begingroup
\renewcommand\thefootnote{*}%
\footnotetext{In this study we categorize action shaping as an action‑space reduction technique.}
\endgroup

Rather than masking actions, some approaches encode them more economically. Majeed and Hutter~\cite{Exact_Reduction} propose action sequentialisation, decomposing each discrete action into a logarithmic-length sequence of binary decisions; this yields a logarithmic reduction in complexity while preserving optimality.  
Tavakoli et al.~\cite{Action_Branching} present the Branching Dueling Q-Network (BDQ), which factorizes a high-dimensional action vector into independent branches sharing a single value module. This causes the network’s output dimensionality to scale linearly—rather than exponentially—with the number of action dimensions and produces strong results on continuous-control benchmarks.

Farquhar \textit{et al.}~\cite{Growing_Action_Spaces} invert the usual pruning narrative through Growing Action Spaces (GAS). Training begins with a small, easily explored subset of actions and gradually unlocks new ones, guided by hierarchical action representations and shared state embeddings. Evaluations on MuJoCo \cite{noauthor_mujoco_nodate} control tasks and multi-agent StarCraft scenarios show substantial sample-efficiency gains by focusing early learning on coarse behaviours before refining fine-grained choices.

\paragraph{Action Masking}

Dynamic masking techniques restore the flexibility that static shaping sacrifices. The BMAS framework~\cite{Learning_State_Specific} computes a bisimulation-based relevance score and suppresses redundant actions per state, complementing global shaping with fine-grained adaptability.  
Action Elimination DQN (AE-DQN)~\cite{Learn_What_Not} employs an auxiliary classifier to predict a state-conditioned elimination signal that discards provably sub-optimal actions, accelerating exploration in combinatorial domains such as the text game Zork \cite{hausknecht_interactive_2020}.Woo and Sung~\cite{Dynamic_Action_Space} go further by continually shrinking the available set during learning: in a $5\times5$ Tic-Tac-Toe environment \cite{brockman_openai_2016} they remove $99.67\,\%$ of all state–action pairs while matching the performance of vanilla Q-learning, demonstrating the power of context-aware elimination.

Collectively, these studies demonstrate that careful management of the action space—whether by static shaping, dynamic masking, compact encoding, or staged expansion—can dramatically reduce both sample complexity and computational load \cite{Action_Space_Shaping}. Nonetheless, most methods presuppose reliable elimination signals or explicit transition–reward structure, and few tackle non-stationary action relevance or cross-task generalisation, leaving ample scope for future research.

\subsection{Action Masking for Autonomous Driving and Mobile Robot Navigation}

Tsampazis et~al.~\cite{tsampazis_deep_2023} study differential-drive navigation with only inexpensive ultrasonic and bumper sensors. They design a simple mask that forbids forward or turning actions when obstacles are sensed, and a sensor-adaptive mask that switches between ultrasonic and touch signals to filter invalid moves more robustly. Using Maskable PPO and a curriculum of procedurally generated Webots maps~\cite{michel2004webotstmprofessionalmobilerobot}, the agent exceeds $98\,\%$ success and remains resilient even when $50\,\%$ of the distance sensors are disabled.

Rudolf et~al.~\cite{rudolf_fuzzy_2022} inject traffic-rule knowledge into a Parameterized DQN by fuzzy-logic masks over the discrete lateral action set (lane changes), while the continuous longitudinal parameters are produced unmasked. Membership functions~\cite{zadeh1996fuzzy} prohibit maneuvers such as right-side overtakes and cutting off faster traffic, accelerating convergence in CommonRoad highway-merging benchmarks~\cite{althoff2017commonroad}.  For unsignalized, mixed-traffic intersections, Guo et~al.~\cite{guo_intersection_2024} propose MSA-PPO: a finely discretized hybrid action space (acceleration, steering) pruned by hard masks that enforce speed limits and ban illegal turns. A self-attention module prioritizes the most influential vehicles, enabling convergence within a few hundred episodes and outperforming a suite of baseline algorithms in reward, success rate, and average speed.

Krasowski et~al.~\cite{krasowski_safe_2020} use a set-based predictor to compute worst-case occupancies of surrounding vehicles. A safety layer filters out high-level lane-change actions that intersect these sets, yielding a dynamic mask for a PPO agent; if no safe action remains, an adaptive-cruise-control fallback is triggered.  
On the highD dataset \cite{krajewski_highd_2018} the framework avoids all collisions and converges substantially faster than unsafe exploration.

Isele et~al.~\cite{isele_safe_2018} employ probabilistic occupancy forecasts of other traffic participants to mask unsafe actions online, restricting exploration to a safe subspace.  
Although the guarantees are probabilistic, the agent learns collision-free intersection policies in stochastic multi-agent settings, illustrating the practicality of prediction-based masking.

Zhang et~al.~\cite{zhang_lexicographic_2023} embed thresholded lexicographic objectives into PPO and apply objective-specific masks in a multi-branch actor–critic, achieving superior stability on MetaDrive \cite{li_metadrive_2023} and SMARTS \cite{zhou_smarts_2021}.  

Bouktif et~al.~\cite{Traffic_Signal_Control} demonstrate that hierarchical parameterized actions—traffic-light phase (discrete) plus duration (continuous)—can be optimized efficiently with P-DQN, underscoring the versatility of hybrid action spaces. Müller et~al.~\cite{muller_safe_2022} formulate the action space as non-conflicting signal phases, encode legal phase transitions in a temporal directed graph, and derive an action mask that guarantees safety a priori.  A secondary psychological mask suppresses erratic phase switches to respect driver comfort. Masked PPO agents converge faster, are more stable, and achieve higher cumulative rewards than unmasked counterparts—advancing RL-based traffic-signal control toward real-world deployment.

Across embodied domains—from mobile robots to autonomous driving and traffic-signal control—invalid-action masking consistently accelerates learning, enforces safety, and improves policy robustness \cite{stolz_excluding_2024, varricchione_pure-past_2024, huang_closer_2022}. 
These works highlight domain knowledge (e.g., traffic rules, sensor affordances) and predictive models as effective sources for constructing masks, while raising open questions about scalable mask inference, probabilistic safety guarantees, and integration with hybrid or multi-objective action spaces.

%Although static shaping, masking, compact encoding and staged expansion have all proven effective at reducing search complexity in generic benchmarks (e.g. StarCraft II, MuJoCo, text games) and even in rudimentary navigation tasks, they typically depend on strong transition–reward knowledge, or simplified dynamics. Moreover, very few methods have been systematically evaluated in a high-fidelity, safety-critical driving simulator, and none have contrasted context-aware dynamic masking against a relative action representation in the same controlled setting. To bridge this gap, we propose and evaluate, within CARLA, four distinct steering action configurations— two full baseline, three fixed discretizations, two state-conditioned dynamic reducer, and two novel relative-steering parameterization—all using the same PPO backbone. By benchmarking convergence speed and control smoothness metrics across diverse scenarios, we demonstrate that adaptive reductions (dynamic and relative) can significantly accelerate learning and improve adaptability without sacrificing precision. This systematic comparison not only highlights the practical value of context-aware action design for autonomous vehicles but also lays the groundwork for future cross-domain generalization of these techniques.

Most action-space reduction/masking studies either evaluate in non-driving simulators—Atari, MuJoCo, ViZDoom, text games—where safety and vehicle kinematics are irrelevant, or they apply fixed, rule-based masks in driving tasks that ignore the agent’s current state. To the best of our knowledge, no prior work jointly benchmarks genuinely state-aware reductions/masking in a high-fidelity self-driving environment. We close this gap by proposing two novel, state-conditioned techniques: (i) dynamic masking, which filters kinematically infeasible steering–throttle pairs online, and (ii) relative reduction, which re-parametrises actions as bounded adjustments around the current steering wheel angle. %Embedded in a multimodal PPO that fuses BEV semantic vision with five scalar vehicle states, these techniques are tested across nine configurations in CARLA. The Relative ±0.5 variant reaches its reward plateau in about 1 M steps— approximately 2 × faster than the full ±1.0 baseline—while maintaining comparable success. These results highlight context-aware action design as a decisive lever for sample-efficient autonomous-driving RL; the next section details our method.

%%%%%%%%%%%%%%%
\section{Method}
This section presents the proposed action space modification strategies and the RL framework used for AV in simulation. Our approach is built upon a discrete action space, where each action corresponds to a specific pair of steering and throttle values. This discrete formulation allows for tractable evaluation of different action space configurations and enables efficient integration of structured reduction techniques. %We begin by describing the baseline action spaces (full and fixed), followed by our proposed dynamic and relative reduction methods. We then detail the multimodal state representation, policy network architecture, and training setup used in the CARLA simulator.

\subsection{Action Space Configurations}
To evaluate the effect of action-space modification, we compared two newly introduced methods—dynamic action-space masking and relative action-space reduction—with two baselines: full action-space and fixed action-space reduction.

\subsubsection{Dynamic Action Space Masking} At each timestep, the steering action space is dynamically reduced to five contextually relevant values centered around the agent’s current steering action as long as the steering is not the two edge or adjacent to two edge values, while throttle actions remain $\{0, 0.2\}$. This configuration combines adaptability with reduced complexity, allowing the agent to focus on meaningful actions at each timestep. For this, we mask the actions that are not valid at each timestep, while still considering the whole action space. This is done because the RL agent expects the same action space at all times, and changing the action space instead of masking invalid actions may confuse it.
To be more precise, the agent observes different outcomes for the same action index, preventing it from learning a consistent mapping between actions and their effects. For instance, by choosing the action 0 when the valid action space is $\{-0.2,-0.1,0\}$ the agent chooses the action -0.2 for steering, while if the valid action space is $\{0.3,0.4,0.5\}$ the agent is choosing the action 0.3 for steering. These actions although considered the same for the agent, has totally different effects in the environment and limit the agent ability to learn the effect of action 0 in the environment. As a result we decided to consider the whole action space and mask the actions that are not valid at each timestep. This way, we eliminated unrealistic exploration of actions in the environment which should lead to faster training. The full action space configurations before masking are:
    \begin{itemize}
    \item Steering actions are discretized between $[-0.5, 0.5]$ with the step of 0.1, throttle $\{0, 0.2\}$.
    \item Steering actions are discretized between $[-1, 1]$ with the step of 0.1 , throttle $\{0, 0.2\}$.
    \end{itemize}
    \paragraph{Computing the Valid Action Space}
    At each timestep $t$, the agent's previous steering index $s_t$ determines the valid steering options for the next step. Here, the $i_f$ denote the full set of steering indices. The valid steering indices for the next step are determined as: 
    \begin{itemize}
    \item If $s_t = 0$ (left boundary), the valid steering options are $\{0,1,2\}$.
    \item If $s_t$ is at the right boundary, the valid options are $\{s_t-2,s_t-1, s_t\}$.
    \item Otherwise, valid steering indices satisfy:
    \begin{equation}
        \{ i \ | \ |i - s_t| \leq 2, \quad i \in i_f \}
    \end{equation}
\end{itemize}

\paragraph{Mask Generation}
Given the valid steering indices, the binary action mask $M_t$ is defined over the flattened action space. The total number of actions depends on the full steering range:

\begin{itemize}
    \item For range $[-0.5, 0.5]$ with 0.1 step: 11 steering values $\Rightarrow |M_t| = 11 \times 2 = 22$
    \item For range $[-1, 1]$ with 0.1 step: 21 steering values $\Rightarrow |M_t| = 21 \times 2 = 42$
\end{itemize}

Let $S$ be the total number of discrete steering actions and $T = 2$ (for throttle options $\{0, 0.2\}$). 

The mask is initialized and filled as follows:

\begin{enumerate}
    \item Initialize $M_t$ as a zero vector of length $S \times T$.
    \item For each valid steering index $s'$ in the valid set:
    \begin{itemize}
        \item For each throttle index $t' \in \{0, 1\}$:
        \begin{itemize}
            \item Compute index = $s' \times T + t'$
            \item Set $M_t[\text{index}] = 1$
        \end{itemize}
    \end{itemize}
\end{enumerate}

This results in a binary mask where only the valid actions for the next timestep are marked as available.

\subsubsection{Relative Action Space Reduction}
    In this method, we consider five different actions for steering $\{-0.2, -0.1,0,0.1, 0.2\}$ and two actions for throttle $\{0,0.2\}$. However, in this method, the steering values are not the exact steering of the vehicle. Instead, the steering actions are relative steering which means that they would say at each timestep, how the RL agent should change the current steering. %It can either go left, got right or stay the same. %This condition is valid as long as the steering is in the valid range of either $[-0.5,0.5]$ or $[-1,1]$.
    The steering value at time step \( t \) is updated based on a discrete steering adjustment:

    \begin{equation}
    s_t = s_{t-1} + \Delta s
    \end{equation}

    where \( s_{t-1} \) is the current steering angle, and \( \Delta s \) is the discrete steering adjustment selected by the RL agent.

The relative steering action is bounded to ensure that the resulting $s_t$ remains within a valid steering range. We consider two ranges: $[-0.5, 0.5]$ and $[-1, 1]$, corresponding to the two variants evaluated in this study.

\paragraph{Action Masking for Invalid Adjustments}
At each timestep, not all relative steering changes are feasible. For instance, if the current steering is near the left or right limit, certain actions would push the resulting value beyond the valid range. To address this, we apply masking to prevent the agent from selecting such infeasible actions during training. Invalid actions are masked as follows:
\begin{itemize}
    \item If the current steering is at or adjacent to the leftmost bound, further negative adjustments are masked.
    \item If the current steering is at or adjacent to the rightmost bound, further positive adjustments are masked.
\end{itemize}

This prevents unrealistic control changes and encourages the agent to learn smooth, physically plausible behaviors. If the agent selects a relative action that would slightly exceed the boundary due to rounding, the final steering value is clipped to the valid range. This method retains a consistent action space size, enabling stable training while dynamically restricting invalid transitions. As with the dynamic masking configuration, this design improves training efficiency and decision stability by avoiding exploration of irrelevant or invalid control commands.
%\end{itemize}

\subsection{Baseline Action Spaces}  % ← one shared heading
To establish meaningful benchmarks for our proposed reduction strategies, we compare against two sets of baseline action spaces. First, the full action spaces discretize both steering and throttle across their entire valid ranges, representing the maximal control granularity. Second, the fixed reduced action spaces constrain the agent to a small, hand-picked set of steering–throttle pairs, reflecting a static pruning approach. The exact range for each is explained below:
\begin{itemize}[wide=0pt, leftmargin=*]

  \item Full Action Space
  \begin{itemize}
    \item Steering discretized in $[-0.5, 0.5]$ with a step of $0.1$; throttle in $\{0,\,0.2\}$.
    \item Steering discretized in $[-1, 1]$ with a step of $0.1$; throttle in $\{0,\,0.2\}$.
  \end{itemize}
  These configurations represent the most complex action spaces and serve as baselines.

  \item Fixed Reduced Action Spaces
  \begin{itemize}
    \item Steering $\{-0.2,\,-0.1,\,0,\,0.1,\,0.2\}$; throttle $\{0,\,0.2\}$.
    \item Steering $\{-0.2,\,0,\,0.2\}$; throttle $\{0,\,0.2\}$.
    \item Steering $\{-0.1,\,0,\,0.1\}$; throttle $\{0,\,0.2\}$.
  \end{itemize}
These configurations serve as baselines against which we compare our proposed methods in terms of convergence speed and driving performance.

\end{itemize}

\begin{comment}
\subsubsection{Full Action Space} 
    \begin{itemize}
    \item Steering actions are discretized between $[-0.5, 0.5]$ with the step of 0.1, throttle $[0, 0.2]$.
    \item Steering actions are discretized between $[-1, 1]$ with the step of 0.1, throttle $[0, 0.2]$.
    \end{itemize}
    These configuration represents the most complex action space and serves as a baselines.
\subsubsection{Fixed Reduced Action Spaces}
    \begin{itemize}
        \item Steering $[-0.2, -0.1, 0, 0.1, 0.2]$, throttle $[0, 0.2]$. 
        \item Steering $[-0.2, 0, 0.2]$, throttle $[0, 0.2]$.
        \item Steering $[-0.1, 0, 0.1]$, throttle $[0, 0.2]$.
    \end{itemize}
    These configurations reduce the number of available actions, simplifying the agent's decision-making process.
\end{comment}

\subsection{Reinforcement Learning Framework}

%\begin{comment}
%\textbf{Algorithm:} 
PPO was selected for its stability, and efficiency making it well-suited for the proposed action space reduction strategies. PPO improves upon traditional policy gradient methods by introducing a clipped surrogate objective that stabilizes updates while maintaining sample efficiency. PPO optimizes the policy by computing the gradient:

%\textbf{Policy Gradient Estimation:}PPO is a policy gradient method that optimizes the policy by computing the gradient:

\begin{equation}
\hat{g} = \mathbb{E}_t \left[ \nabla_\theta \log \pi_\theta(a_t | s_t) \hat{A}_t \right],
\end{equation}
where \(\pi_\theta(a_t | s_t)\) is the policy parameterized by \(\theta\), \(\hat{A}_t\) is the advantage estimate at timestep \(t\), and the expectation is computed over a batch of trajectories. To ensure stable updates, PPO introduces a clipped surrogate objective:

%\textbf{Clipped Surrogate Objective:} To ensure stable updates, PPO introduces a clipped surrogate objective:

\begin{equation}
L_{\text{CLIP}}(\theta) = \mathbb{E}_t \left[ \min \left( r_t(\theta) \hat{A}_t, \text{clip}(r_t(\theta), 1 - \epsilon, 1 + \epsilon) \hat{A}_t \right) \right],
\end{equation}

where \(r_t(\theta) = \frac{\pi_\theta(a_t | s_t)}{\pi_{\theta_\text{old}}(a_t | s_t)}\) is the probability ratio, and \(\epsilon\) is a hyperparameter controlling the extent of clipping. This objective prevents overly large policy updates by clipping \(r_t(\theta)\) within \([1 - \epsilon, 1 + \epsilon]\) when the advantage \(\hat{A}_t\) would otherwise encourage extreme changes. By taking the minimum of the clipped and unclipped terms, \(L_{\text{CLIP}}\) provides a pessimistic bound on the policy improvement.

%\textbf{Advantage Function:}  
The advantage function \(\hat{A}_t\) is a critical component in PPO as it quantifies how much better or worse an action \(a_t\) is compared to the average action the policy would take in the state \(s_t\). Formally, it is defined as:
\begin{equation}
\hat{A}_t = Q(s_t, a_t) - V(s_t),
\end{equation}
where \(Q(s_t, a_t)\) represents the expected return (cumulative discounted reward) from taking action \(a_t\) in state \(s_t\), and \(V(s_t)\) is the value function estimating the expected return from state \(s_t\) regardless of the action.

In practice, PPO uses Generalized Advantage Estimation (GAE) to compute \(\hat{A}_t\), which balances bias and variance by incorporating a decay factor \(\lambda\):
\begin{equation}
\hat{A}_t = \delta_t + (\gamma \lambda) \delta_{t+1} + \cdots + (\gamma \lambda)^{T-t-1} \delta_{T-1},
\end{equation}
where:
\begin{itemize}
    \item \(T\) is the trajectory segment length,
    \item \(\delta_t\), the temporal-difference (TD) error,
    \item \(\gamma\) is the discount factor, controlling the weight of future rewards.
\end{itemize}
The temporal-difference (TD) error \(\delta_t\), is given by:
\begin{equation}
\delta_t = r_t + \gamma V(s_{t+1}) - V(s_t).
\end{equation}
Here, \(\gamma\) is the discount factor controlling the importance of future rewards, and \(\lambda\) controls the tradeoff between using high-bias, low-variance single-step estimates and low-bias, high-variance multi-step estimates. The advantage function enables the algorithm to prioritize actions that perform better than expected and reduce the probability of actions that underperform.

%\textbf{Combined Objective:} 
The PPO objective incorporates the value loss and entropy regularization to enhance training stability and ensure sufficient exploration:
\begin{equation}
L(\theta) = \mathbb{E}_t \left[ L_{\text{CLIP}}(\theta) - c_1 L_V(\theta) + c_2 S[\pi_\theta](s_t) \right],
\end{equation}
where \(L_V(\theta)\) is the squared error between the predicted value \(V_\theta(s_t)\) and the target return, \(S[\pi_\theta](s_t)\) represents the entropy of the policy for encouraging exploration, and \(c_1, c_2\) are coefficients controlling the weight of these terms.

%\textbf{Algorithm Steps:} 

The PPO algorithm alternates between data collection and policy optimization which is shown in the algorithm \ref{alg:ppo-simple}.

%\end{comment}

\begin{comment}
\begin{enumerate}
    \item Collect trajectories by running the current policy \(\pi_\theta\) in the environment.
    \item Compute advantage estimates \(\hat{A}_t\) using GAE.
    \item Optimize the combined objective \(L(\theta)\) for \(K\) epochs using minibatches of data.
    \item Update the policy parameters: \(\theta \leftarrow \theta_{\text{new}}\).
\end{enumerate}
\end{comment}

%\begin{comment}
\begin{algorithm}[H]
\caption{PPO: Data Collection and Policy Optimization}\label{alg:ppo-simple}
\begin{algorithmic}[1]
  \REQUIRE initial policy parameters $\theta_0$, number of iterations $T$, epochs per iteration $K$
  \FOR{$i = 0,1,\dots,T-1$}
    \STATE \textbf{(1) Collect trajectories:} run $\pi_{\theta_i}$ in env.\ for $N$ steps, storing $(s_t,a_t,r_t)$
    \STATE \textbf{(2) Compute advantages:} $\hat A_t \gets \text{GAE}(r_t, V_{\theta_i})$
        \STATE \textbf{(3) Policy update:} for $K$ epochs, optimize
  \[
    L(\theta) = \mathbb{E}_t\Bigl[L_{\mathrm{CLIP}}(\theta) - c_1 L_V(\theta) + c_2 S[\pi_\theta](s_t)\Bigr]
  \]

      using minibatches
    \STATE \textbf{(4) Set} $\theta_{i+1} \gets \theta$
  \ENDFOR
  \ENSURE final policy $\pi_{\theta_T}$
\end{algorithmic}
\end{algorithm}
%\end{comment}

\begin{comment}
PPO offers several advantages for this study:
\begin{itemize}
    \item Stability: The clipping mechanism ensures stable and reliable updates, critical when testing different action space configurations.
    \item Sample Efficiency: By reusing data for multiple epochs of optimization, PPO reduces the sample complexity, making it cost-effective for high-fidelity environments like CARLA.
    \item Exploration and Robustness: The entropy regularization promotes exploration, preventing premature convergence to suboptimal policies.
\end{itemize}
\end{comment}

PPO brings several key benefits to this study. First, its clipping mechanism helps maintain stability by preventing overly large policy updates, which is especially important when evaluating various action‐space configurations. Second, PPO’s ability to reuse collected data across multiple optimization epochs improves sample efficiency, making it more cost‐effective to train in a computationally intensive, high‐fidelity environment like CARLA.

At each step the agent receives an observation, chooses an
action, and collects a reward until an episode ends
under the termination signal.  Concretely,
\begin{itemize}
  \item State (observation) combines a stack of four BEV frames
        with five scalar features (throttle, speed, steering, lane offset,
        heading angle).
  \item Reward encourages centred, collision-free, progressive
        driving and penalises overspeed and lane departures.
  \item Termination is triggered by goal reach, collision,
        time-out, prolonged low speed, excessive offset, overpass the goal, or overspeed.
\end{itemize}
Full formulas and thresholds appear in
Section~\ref{sec:env}.

\section{Experimental setup} \label{sec:env}

%\subsection{Environment Description}
Experiments are conducted in the CARLA simulator. We adopt CARLA 0.9.15 in synchronous mode. The agent is trained and evaluated in Town07, which features a diverse set of road layouts, including straight roads, intersections, and curves, providing a suitable testbed for RL-based driving policies. Our choice of Town07 and the high-level reward structure and the state space were partly inspired by the open-source study of Shaikh
et al.~\cite{shaikh2023carla}; however, all code was re-implemented
from scratch in Python using Stable-Baselines3 \cite{stable-baselines3}.

\subsection{Training Procedure}
\begin{figure}[!t]
    \centering
    \includegraphics[width=0.5\textwidth]{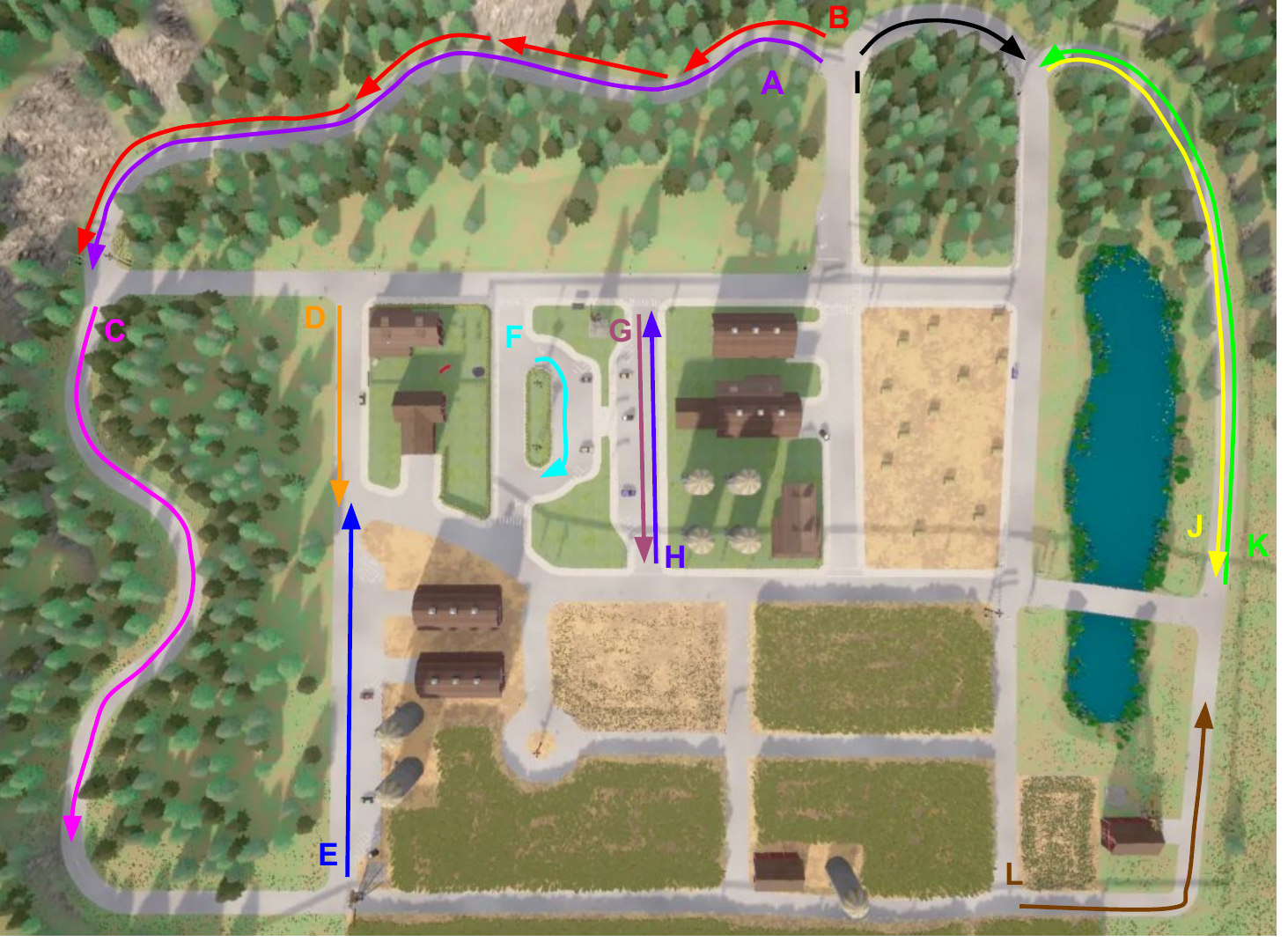}  % Adjust width as needed
    \caption{Routes in CARLA's Town07 used for training and testing. Route A represents the whole training path. Routes B are the 4 different parts the training route was divided for training. Routes C and F are multi-turn routes in test scenarios, routes I, J, K, L show one-turn routes and routes D, E, G, H shows the straight routes for test scenarios.}
    \label{fig:tracks_town07}
% \vspace*{-10pt}
\end{figure}

The RL agent is trained using PPO. The masking for different action space was done using MaskablePPO in Stable-Baselines3. The training was done on a long route in Town07 that was splited into four different sections to prevent overfitting to the route. Fig. \ref{fig:tracks_town07} shows the whole route and each section inside it. The agent is getting trained by randomly selecting each of these four sections at each timestep. The policy network receives both visual and non-visual inputs and is optimized using the following hyperparameters:

\begin{itemize}
    \item Learning Rate: \(3 \times 10^{-4}\) with linear decay.
    \item Discount Factor (\(\gamma\)): \(0.99\).
    \item Generalized Advantage Estimation (GAE) Parameter (\(\lambda\)): \(0.95\).
    \item Batch Size: \(256\).
    \item Rollout Length: \(8192\) environment steps per policy update.
    \item Epochs per Update: \(15\).
    \item PPO Clipping Range: \(0.2\).
    \item Entropy Coefficient: \(0.01\).
    \item Value Function Coefficient: \(0.5\).
\end{itemize}

To process observations efficiently, we employ a MobileNetV3-based feature extractor for image encoding, followed by a Conv1D module for temporal feature aggregation. The model architecture for the autoencoder is shown in Table \ref{tab:autoencoder}. The encoder part was used as feature extractor used in RL training. The non-visual numerical features are encoded using a fully connected network and fused with the visual embeddings before passing them to the policy and value networks.

\begin{table}[h]
    \centering
    \caption{MobileNetV3-Based Autoencoder Architecture}
    \label{tab:autoencoder}
    \begin{tabular}{l c c c }
        \hline
        \textbf{Layer Type/ Block} & \textbf{Kernel / Units} & \textbf{Size} & \textbf{Activation} \\
        \hline
        \hline
        \multicolumn{4}{c}{\textbf{Encoder}} \\
        \hline
        \hline
        Input & - & (64, 64, 3) & - \\
        \hline
        Conv2D + BN & 3×3, stride 2 & 16 filters & ReLU \\
        \hline
        MobileNetV3 Blocks & - & 80 filters & ReLU \\
        \hline
        Flatten & - & 1280 & - \\
        \hline
        Dense (FC) & 512 & 512 & ReLU \\
        \hline
        \hline
        \multicolumn{4}{c}{\textbf{Decoder}} \\
        \hline
        \hline
        Dense (FC) & 512 $\rightarrow$ 1280 & (4×4×80) & ReLU \\
        \hline
        Unflatten & - & (80, 4, 4) & - \\
        \hline
        ConvTranspose2D + BN & 3×3, stride 2 & 64 filters & ReLU \\
        \hline
        ConvTranspose2D + BN & 3×3, stride 2 & 32 filters & ReLU \\
        \hline
        ConvTranspose2D + BN & 3×3, stride 2 & 16 filters & ReLU \\
        \hline
        ConvTranspose2D & 3×3, stride 2 & 3 filters & Sigmoid \\
        \hline
    \end{tabular}
\end{table}

%\subsubsection{Evaluation Methodology}
%To assess the generalization capability of the learned policy, we evaluate the trained agent on unseen test scenarios in Town07. These test scenarios include the complete training route and 10 novel trajectories, ensuring a comprehensive assessment of the agent’s adaptability.
\subsubsection{State Space}

The agent's state space is designed to integrate visual and scalar data, enabling the RL agent to perceive the environment comprehensively. It is defined as a dictionary of observations, normalized for consistent input scaling. The observations are preprocessed and normalized before being passed to the agent, ensuring consistent feature scaling. Each component of the state space is explained below:

\begin{itemize}
    \item \textit{Image Data:} The key visual input consists of a stack of four consecutive frames, normalized to the range \([0, 1]\), with dimensions. Each frame is captured at regular intervals, providing temporal context for motion or environmental changes.
    \item \textit{Throttle:} The throttle input, represented as a scalar, indicates the current acceleration level. It is normalized to the range \([0, 1]\) for consistency.
    \item \textit{Velocity:} The agent's velocity is normalized by dividing by a maximum velocity, producing values in the range \([0, 1]\). This allows the agent to reason about relative speeds.
    \item \textit{Current Steer:} The current steering angle is represented as a scalar in the range \([-1, 1]\), where \(-1\) and \(1\) correspond to maximum left and right steering, respectively.
    \item \textit{Distance from Center:} The distance of the agent from the lane center is normalized by a maximum distance (e.g., 3 meters), ensuring values fall within \([0, 1]\). This provides critical feedback for lane-keeping tasks.
    \item \textit{Angle:} The angle between the vehicle's heading and the desired direction is normalized by \(\pi\), producing values in \([0, 1]\). This feature helps the agent align itself with the correct trajectory.
\end{itemize}

To encode the multi-modal observation space, we implement a custom feature extractor that processes both image-based and scalar inputs. The network comprises three main components:

\begin{itemize}
    \item \textit{Visual Feature Processing:}
    \begin{itemize}
        \item The input image consists of four consecutive RGB frames, stacked along the channel dimension and reshaped appropriately to preserve temporal order.
        \item Each frame is passed individually through a pretrained MobileNetV3-based encoder, yielding a 512-dimensional latent vector per frame.
        \item These frame-wise embeddings are stacked and processed via a temporal \texttt{Conv1D} module composed of two convolutional layers with ReLU activations and an adaptive average pooling layer. This module captures temporal dependencies and outputs a 128-dimensional aggregated feature vector.
    \end{itemize}
    
    \item \textit{Scalar Feature Processing:}
    \begin{itemize}
        \item Five scalar inputs—throttle, velocity, current steer, lateral deviation from lane center, and angle to waypoint—are each processed through an independent feedforward sub-network with one hidden layer and ReLU activation.
        \item Each scalar is embedded into a 32-dimensional vector, and the five embeddings are concatenated and passed through an additional fusion module to produce a 32-dimensional combined scalar representation.
    \end{itemize}
    
    \item\textit{Feature Combination:}
    \begin{itemize}
        \item The final feature vector is obtained by concatenating the 128-dimensional visual-temporal feature vector and the 32-dimensional scalar representation, resulting in a 160-dimensional fused representation.
        \item This combined vector is further processed by a fully connected layer with ReLU activation to produce the final 160-dimensional feature vector used by both the policy and value networks.
    \end{itemize}
\end{itemize}

\subsubsection{Reward Function}

The reward function is designed to guide the agent towards safe and efficient driving behavior by incorporating multiple reward components that evaluate key
aspects of vehicle performance. The reward function considers lane centering, speed management, collision avoidance, and progress toward the goal. The individual components of the reward function are detailed below:

\begin{itemize}
    \item \textit{Lane Centering Reward:}
    The reward encourages the agent to stay close to the center of the driving lane. If the lateral lane offset is within 1.5 meters, the reward is computed as:
    \begin{equation}
    r_{\text{lane}} = \max(0, 1.5 - \text{lane\_offset}) \times 40.
    \end{equation}

    \item \textit{Heading Alignment Reward:}
    The agent is rewarded for aligning its heading with the road direction. If the angular deviation from the road’s heading is within 0.2 (normalized), the reward is:
    \begin{equation}
    r_{\text{angle}} = \max(0, 0.2 - \text{angle\_offset}) \times 100.
    \end{equation}
\begin{comment}
    \item \textit{Speed Management:}  
    \begin{itemize}
        \item If the vehicle’s speed is between 1 and 25 km/h, a fixed reward of 10 is added.
        \item If the vehicle speed exceeds 25 km/h, a penalty is applied:
        \begin{equation}
        r_{\text{overspeed}} = -2 \times (\text{speed} - 25).
        \end{equation}
        \item If the vehicle speed remains below 1 km/h:
        \begin{itemize}
            \item For less than 10 seconds, a penalty is incrementally applied:
            \begin{equation}
            r_{\text{low-speed}} = -2.0 \times \text{elapsed\_time}.
            \end{equation}
            \item For 10 seconds and above, a fixed penalty of -50 is applied and overrides all other rewards. This will also leads to termination of the episode.
        \end{itemize}
    \end{itemize}
\end{comment}  

\item \textit{Speed Management:} The reward is computed as:

\[
r_{\text{speed}} =
\left\{
\begin{array}{ll}
+10, & \text{if } 1 \leq v \leq 25 \\
-2 \times (v - 25), & \text{if } v > 25 \\
-2 \times t, & \text{if } v < 1 \text{ and } t < 10 \\
-50, & \text{if } v < 1 \text{ and } t \geq 10
\end{array}
\right.
\]

Here, $v$ is the vehicle's speed in km/h, and $t$ is the time (in seconds) the vehicle has been moving slower than $1$ km/h.  
The episode terminates if the last condition is met.

    \item \textit{Lane Invasion Penalty:}  
    Each lane invasion is penalized proportionally:
    \begin{equation}
    r_{\text{invasion}} = -\frac{\text{total\_lane\_invasions}}{4}.
    \end{equation}

    \item \textit{Progress Toward the Goal:}  
    The agent receives a scaled reward based on how much distance it has covered toward the goal:
    \begin{equation}
    r_{\text{goal-progress}} = \max \left(0, \frac{\text{init\_goal\_dist} - \text{goal\_dist}}{\text{init\_goal\_dist}} \right) \times 60.
    \end{equation}
    Additionally, a high reward of +200 is provided if the vehicle is within 2 meters of the goal.

    \item \textit{Collision Penalty:}  
    If the vehicle collides during the episode, the total reward is overwritten with a penalty of:
    \begin{equation}
    r_{\text{collision}} = -50.
    \end{equation}
\end{itemize}

The total reward is primarily computed as the sum of all individual components:
\begin{multline}
r_{\text{total}} = r_{\text{lane}} + r_{\text{angle}} + r_{\text{speed}} \\
+ r_{\text{goal-progress}} + r_{\text{invasion}} + r_{\text{collision}}.
\end{multline}

However, in critical cases such as collision events or prolonged low-speed behavior (exceeding 10 seconds below 1 km/h), the total reward is overridden and set to the corresponding penalty value. This mechanism ensures that dangerous or undesired behaviors are strictly discouraged, regardless of positive contributions from other components.

\subsubsection{Termination Condition}

The episode is terminated based on several conditions that ensure the agent adheres to safe and goal-directed driving behavior. These termination conditions are designed to handle cases such as reaching the goal, collisions, or unsafe driving practices. The following criteria are used to determine if an episode should end:

\begin{itemize}
    \item \textit{Goal Achievement:}  
    The episode terminates when the vehicle is within \(2.0 \, \text{meters}\) of the goal location. This ensures that the agent has successfully completed its task.
    
    \item \textit{Collision Detection:}  
    The episode ends if the vehicle detects any collision during the run. Collisions are tracked using a collision history buffer, and any non-zero collisions result in termination.

    \item \textit{Time Limit:}  
    The episode terminates if the elapsed time exceeds a predefined limit, \(\text{SECONDS\_PER\_EPISODE}\) which is 600 seconds for this study. This ensures that the training process does not stagnate due to excessively long episodes.

    \item \textit{Passed Goal}: 
    If the vehicle begins moving away from the goal after having reached its minimum distance (tracked over time), and fails to improve proximity for more than 500 steps, the episode is terminated. This prevents inefficient circling or missed-goal scenarios.

    \item \textit{Prolonged Low Speed:}  
    If the vehicle remains at a speed below \(1.0 \, \text{km/h}\) for more than \(10 \, \text{seconds}\), the episode is terminated. This prevents the agent from idling or failing to start moving.

\end{itemize}
\subsection{Test Scenarios}
To assess the generalization capability of the learned policy, we evaluate the trained agent on unseen test scenarios in Town07. An overview of the full training and testing routes is illustrated in Fig.~\ref{fig:tracks_town07}, where each route is labeled with its corresponding identifier (e.g., A, B, C, etc.). The scenarios include the complete training route and 10 novel trajectories, ensuring a comprehensive assessment of the agent’s adaptability. The 11 distinct test routes categorized into three main track types, namely Straight (D, E, G, H), One-Turn (I, J, K, L), Multiple-turns (C, F), and Full Route (A). The full route (A) represents the complete trajectory used during training, which was divided into four sections for training. During evaluation, the agent was tested on the full trajectory to assess how well it generalized across all road types. Each test scenario is executed for five independent episodes, resulting in a total of 55 test episodes. %test tracks are shown in Fig. \ref{fig:tracks_town07}.

\subsubsection{Evaluation Metrics}
We use the following quantitative metrics to evaluate performance:
\begin{itemize}
    \item \textit{Success Rate:} The percentage of episodes where the agent successfully reaches the goal.
    \item \textit{Episode Reward:} The total accumulated rewards, reflecting the agent's ability to balance speed, lane-keeping, and collision avoidance. 
    \item \textit{Episode Length:} The number of timesteps before termination.
    \item \textit{Lane Deviation:} The average lateral distance from the lane center.
    \item \textit{Efficiency:} The combinaion of the convergence rate and the success rate as explained in \ref{sec:cov}
    
\end{itemize}

All evaluations are conducted deterministically, where the agent selects the action with the highest probability to ensure consistent benchmarking. %To support stable policy learning and effective decision-making, the agent receives a structured state representation that captures both its surroundings and its internal dynamics.% In the following sections, we describe the composition of the observation space, the design of the reward function, and the termination conditions used during training.

\subsubsection{Convergence Rate Computation} \label{sec:cov}
To quantify sample efficiency, we measure each method’s \emph{empirical convergence rate} as the reciprocal of the number of training steps required to reach a fixed performance threshold.  Specifically, let $R_i(t)$ denote the mean episode reward of method $i$ at step $t$, and define
\[
R_{\max} = \max_{i}\max_{0 \le t \le T_{\max}} R_i(t)
\]
as the highest reward attained by any method within a budget of $T_{\max}$ steps.  We then set the target level
\[
R_{\mathrm{tgt}} = p\,R_{\max},\quad p\in (0,1)\ (\text{we use }p=0.6).
\]
For each method $i$, the \emph{step‐to‐target} $T_i$ is the first step at which
\[
R_i(T_i)\;\ge\;R_{\mathrm{tgt}},
\]
%or $T_i=\infty$ if this never occurs.
The empirical convergence rate is then defined as
\[
\mathrm{rate}_i \;=\;\frac{1}{T_i}.
\]
%Methods that do not reach $R_{\mathrm{tgt}}$ within $T_{\max}$ are assigned $\mathrm{rate}_i=0$. We use this step-to-threshold to evaluate which method converges faster. Moreover, we combine this convergence with the success rate and define an efficiency metric that considers both the success rate and the convergence rate to jointly capture reliability and speed of learning as

Methods that fail to reach $R_{\mathrm{tgt}}$ within $T_{\max}$ are assigned a convergence rate of zero. We use this \emph{step-to-threshold} metric to compare how quickly each method converges. Furthermore, we integrate convergence rate with success rate to define an efficiency metric that jointly captures learning reliability and speed:

\[
E_i \;=\;10^{-7}\;\times\;\frac{\mathrm{SR}_i}{100}\;\times\;\frac{1}{T_i}.
\]

\section{Results}
\subsection{Training Performance}

\begin{comment}
\begin{figure}[h]
    \centering
    \includegraphics[width=\linewidth, keepaspectratio]{figs/reward_comparison_up_to_4M.pdf}
    \caption{Mean episode reward during training over 4 million steps for different action space configurations. The full action space with range $[-0.5, 0.5]$ and the relative action space with the same range show the best performance, while broader ranges or fewer discrete actions tend to hinder learning progress.}
    \label{fig:training_reward}
\end{figure}
\end{comment}
\begin{figure*}[ht]
    \centering
    \includegraphics[width=\textwidth, keepaspectratio]{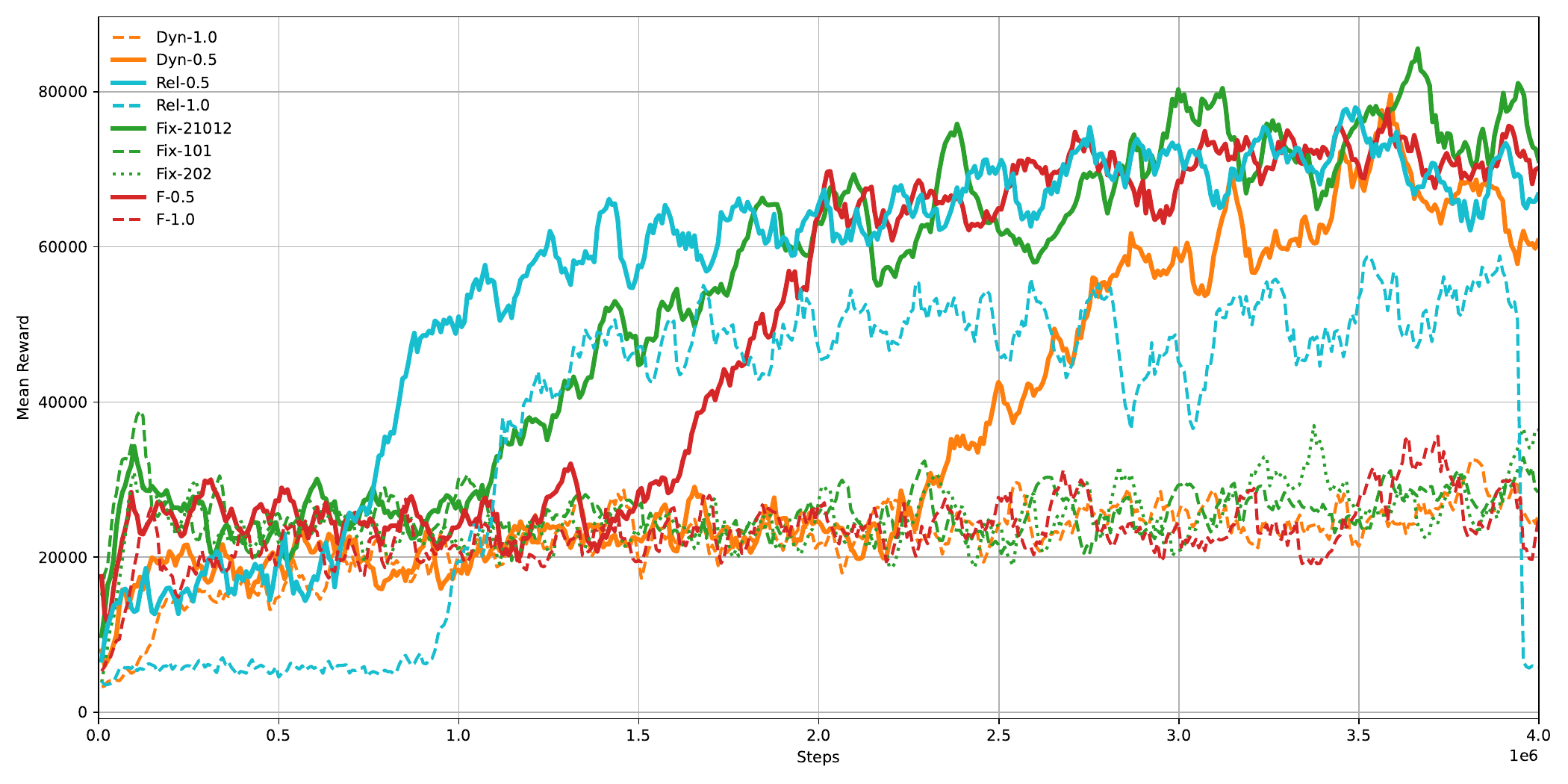}
    \caption{Mean episode reward during training over 4 million steps for different action space configurations.}
    \label{fig:training_reward}
\end{figure*}

This section presents both training performance and route-type-based evaluations of different action space configurations in the context of RL for autonomous driving. To evaluate the impact of different action space reductions, we compare nine representative configurations described in Table \ref{tab:action_space_labels}.
%: Fixed action space reduction 1 $[-0.2, -0.1, 0, 0.1, 0.2]$, Fixed action space reduction 2 $[-0.1, 0, 0.1]$, Fixed action space reduction 3 $[-0.2, 0, 0.2]$ , Full action space $[-0.5, 0.5]$, Full action space $[-1, 1]$, Dynamic action space reduction with action range $[-1, 1]$, Dynamic action space reduction with action range $[-0.5, 0.5]$ , Relative action space reduction with range $[-0.5, 0.5]$, and Relative action space reduction with range $[-1, 1]$.

\begin{table}[htbp]
%\small
\scriptsize
\centering
\caption{Action Space Configurations and Abbreviated Labels}
\label{tab:action_space_labels}
%\begin{tabularx}{\columnwidth}{l l X}
\begin{adjustbox}{max width=\columnwidth}

\begin{tabular}{p{1.2cm} p{4cm} p{4cm}}
\toprule
\textbf{Label} & \textbf{Steering Space} & \textbf{Description} \\
\midrule
F-0.5 & $[-0.5, 0.5]$ (step 0.1) & Full discrete action space \\ 
\addlinespace

F-1.0 & $[-1, 1]$ (step 0.1) & Full discrete action space with (wider range) \\
\addlinespace

Fix-101 & $\{-0.1, 0, 0.1\}$ & Fixed reduction with 3 actions \\
\addlinespace

Fix-202 & $\{-0.2, 0, 0.2\}$ & Fixed reduction with 3 actions (wider range) \\
\addlinespace

Fix-21012 & $\{-0.2, -0.1, 0, 0.1, 0.2\}$ & Fixed reduction with 5 actions (denser coverage) \\
\addlinespace

Dyn-0.5 & Contextual subset of $[-0.5, 0.5]$ & Dynamic masking based on current steering (±0.2) \\
\addlinespace

Dyn-1.0 & Contextual subset of $[-1, 1]$ & Dynamic masking with wider range \\
\addlinespace

Rel-1.0 & $\Delta \in [-0.2, 0.2]$ (with step 0.1 from current steer) & Relative steering within ±0.2 on a $[-1, 1]$ range \\
\addlinespace

Rel-0.5 & $\Delta \in [-0.2, 0.2]$ (with step 0.1 from current steer) & Relative steering within ±0.2 on a $[-0.5, 0.5]$ range \\

\bottomrule
%\end{tabularx}
\end{tabular}
\end{adjustbox}
\end{table}

Fig.~\ref{fig:training_reward} presents the evolution of mean episode reward over 4 million training steps for all nine action space configurations. Among these, the Fix-21012 and the Rel-0.5 demonstrate both fast convergence and high final reward. The Dyn-0.5 and the F-0.5 also show competitive learning behavior with smooth and stable reward growth. 

In contrast, broader range configurations such as F-1.0, Rel-1.0, and Dyn-1.0 result in lower final rewards and greater instability. This behavior can be attributed to either over-exploration or imprecise control granularity, which may hinder fine-tuned lane adjustments and consistent trajectory planning. The narrower fixed setup, Fix-101, Fix-202, also underperforms, likely due to its overly constrained maneuverability.

Based on these observations, we selected the five most promising configurations for further evaluation in the test environment: F-0.5, Dyn-0.5, Fix-21012, Rel-0.5, and Rel-1.0. These configurations offer a strong balance between learning performance and action diversity, and are analyzed in detail in the next section.

\subsection{Evaluation Across Track Types}

\begin{comment}
\begin{table*}[ht]
\centering
\caption{Average Reward and Success Rate Across Track Types}
\label{tab:reward_sr_route_type}
\setlength{\tabcolsep}{4pt}
\begin{tabular}{l|cc|cc|cc|cc|cc}
\hline \hline
\textbf{Track Type} & \multicolumn{2}{c|}{\textbf{Fixed [-0.2,...]}} & \multicolumn{2}{c|}{\textbf{Full [-0.5,0.5]}} & \multicolumn{2}{c|}{\textbf{Dynamic}} & \multicolumn{2}{c|}{\textbf{Rel [-0.5,0.5]}} & \multicolumn{2}{c}{\textbf{Rel [-1,1]}} \\ 
 & \textbf{Reward ↑} & \textbf{SR (\%) ↑} & \textbf{Reward ↑} & \textbf{SR (\%) ↑} & \textbf{Reward ↑} & \textbf{SR (\%) ↑} & \textbf{Reward ↑} & \textbf{SR (\%) ↑} & \textbf{Reward ↑} & \textbf{SR (\%) ↑} \\
\hline
Multi-Turn (C,F)     & 42210.05 & \textbf{50.00} & \textbf{48458.43} & 50.00 & 43647.77 & \textbf{50.00} & 41614.29 & \textbf{50.00} & 37243.74 & 0.00 \\ \hline
One-Turn (I,J,K,L)   & 51047.36 & 50.00 & 45105.48 & \textbf{75.00} & \textbf{53974.43} & 25.00 & 50941.50 & 70.00 & 30757.75 & 50.00 \\ \hline
Straight (D,E,G,H)   & 21216.74 & 75.00 & 16806.96 & 50.00 & 23029.84 & \textbf{100.00} & 23651.53 & \textbf{100.00} & \textbf{25996.38} & 40.00 \\ \hline
Full Route (A)       & 129791.60 & \textbf{100.00} & \textbf{137167.49} & \textbf{100.00} & 130143.41 & \textbf{100.00} & 118056.87 & 80.00 & 109616.06 & 40.00 \\ \hline
\hline
\end{tabular}
\end{table*}
\end{comment}

\begin{table*}[ht]
\centering
\caption{Average Reward and Success Rate Across Track Types}
\label{tab:reward_sr_route_type}
\setlength{\tabcolsep}{4pt}
\begin{tabular}{l|ccc|ccc|ccc|ccc}
\hline \hline
\textbf{Configuration} 
& \multicolumn{3}{c|}{\textbf{Multi-Turn (C,F)}} 
& \multicolumn{3}{c|}{\textbf{One-Turn (I,J,K,L)}} 
& \multicolumn{3}{c|}{\textbf{Straight (D,E,G,H)}} 
& \multicolumn{3}{c}{\textbf{Full Route (A)}} \\

& \textbf{Reward ↑} & \textbf{SR (\%) ↑}  & \textbf{Eff↑}
& \textbf{Reward ↑} & \textbf{SR (\%) ↑} & \textbf{Eff↑}
& \textbf{Reward ↑} & \textbf{SR (\%) ↑} & \textbf{Eff↑}
& \textbf{Reward ↑} & \textbf{SR (\%) ↑} & \textbf{Eff↑}\\

\hline
\textbf{Fix-21012}     
& 42210.05 & \textbf{50.00} & 3.55
& 51047.36 & 50.00 & 3.55
& 21216.74 & 75.00 & 5.32
& 129791.60 & \textbf{100.00} & 7.10\\

\textbf{F-0.5}      
& \textbf{48458.43} &  \textbf{50.00} & 2.65
& 45105.48 & \textbf{75.00} & 3.98
& 16806.96 & 50.00 & 2.65
& \textbf{137167.49} & \textbf{100.00} & 5.30\\

\textbf{Dyn-0.5}              
& 43647.77 & \textbf{50.00} & 1.81
& \textbf{53974.43} & 25.00 & 0.91
& 23029.84 & \textbf{100.00} & 3.63
& 130143.41 & \textbf{100.00} & 3.63 \\

\textbf{Rel-0.5}       
& 41614.29 & \textbf{50.00} & \textbf{4.88}
& 50941.50 & 70.00 & \textbf{6.84}
& 23651.53 & \textbf{100.00} & \textbf{9.77}
& 118056.87 & 80.00 & \textbf{7.81} \\

\textbf{Rel-1.0}           
& 37243.74 & 0.00 & 0.00
& 30757.75 & 50.00 & 3.01
& \textbf{25996.38} & 40.00 & 2.41
& 109616.06 & 40.00 & 2.41 \\

\hline \hline
\end{tabular}
\end{table*}

\begin{table*}[ht]
\centering
\caption{Average Episode Length and Lane Deviation Across Track Types}
\label{tab:length_deviation_route_type}
\setlength{\tabcolsep}{4pt}
\begin{tabular}{l|cc|cc|cc|cc}
\hline \hline
\textbf{Configuration} 
& \multicolumn{2}{c|}{\textbf{Multi-Turn (C,F)}} 
& \multicolumn{2}{c|}{\textbf{One-Turn (I,J,K,L)}} 
& \multicolumn{2}{c|}{\textbf{Straight (D,E,G,H)}} 
& \multicolumn{2}{c}{\textbf{Full Route (A)}} \\

& \textbf{Ep. Len ↑} & \textbf{Lane Dev. ↓} 
& \textbf{Ep. Len ↑} & \textbf{Lane Dev. ↓} 
& \textbf{Ep. Len ↑} & \textbf{Lane Dev. ↓} 
& \textbf{Ep. Len ↑} & \textbf{Lane Dev. ↓} \\

\hline
\textbf{Fix-21012}     
& 478.40 & \textbf{0.17} 
& 527.00 & 0.12 
& 254.80 & 0.18 
& 1361.00 & 0.11 \\

\textbf{F-0.5}      
& 645.30 & 0.31 
& 464.25 & 0.09 
& 219.80 & 0.28 
& 1391.80 & \textbf{0.08} \\

\textbf{Dyn-0.5}              
& 474.00 & 0.19 
& 554.10 & \textbf{0.07} 
& 250.40 & \textbf{0.12} 
& 1294.80 & \textbf{0.08} \\

\textbf{Rel-0.5}       
& 481.00 & 0.24 
& \textbf{570.50} & 0.15 
& 282.60 & 0.17 
& 1278.40 & 0.10 \\

\textbf{Rel-1.0}           
& \textbf{657.20} & 0.52 
& 471.60 & 0.21 
& \textbf{417.20} & 0.40 
& \textbf{1462.60} & 0.17 \\

\hline \hline
\end{tabular}
\end{table*}

\begin{comment}
\begin{table*}[ht]
\centering
\caption{Average Episode Length and Lane Deviation Across Track Types}
\label{tab:length_deviation_route_type}
\setlength{\tabcolsep}{4pt}
\begin{tabular}{l|cc|cc|cc|cc|cc}
\hline \hline
\textbf{Track Type} & \multicolumn{2}{c|}{\textbf{Fixed [-0.2,...]}} & \multicolumn{2}{c|}{\textbf{Full [-0.5,0.5]}} & \multicolumn{2}{c|}{\textbf{Dynamic}} & \multicolumn{2}{c|}{\textbf{Rel [-0.5,0.5]}} & \multicolumn{2}{c}{\textbf{Rel [-1,1]}} \\
 & \textbf{Ep. Len ↑} & \textbf{Lane Dev. ↓} & \textbf{Ep. Len ↑} & \textbf{Lane Dev. ↓} & \textbf{Ep. Len ↑} & \textbf{Lane Dev. ↓} & \textbf{Ep. Len ↑} & \textbf{Lane Dev. ↓} & \textbf{Ep. Len ↑} & \textbf{Lane Dev. ↓} \\
\hline
Multi-Turn (C,F)     & 478.40 & \textbf{0.17} & 645.30 & 0.31 & 474.00 & 0.19 & 481.00 & 0.24 & \textbf{657.20} & 0.52 \\ \hline
One-Turn (I,J,K,L)   & 527.00 & 0.12 & 464.25 & 0.09 & 554.10 & \textbf{0.07} & \textbf{570.50} & 0.15 & 471.60 & 0.21 \\ \hline
Straight (D,E,G,H)   & 254.80 & 0.18 & 219.80 & 0.28 & 250.40 & \textbf{0.12} & 282.60 & 0.17 & \textbf{417.20} & 0.40 \\ \hline
Full Route (A)       & 1361.00 & 0.11 & 1391.80 & \textbf{0.08} & 1294.80 & \textbf{0.08} & 1278.40 & 0.10 & \textbf{1462.60} & 0.17 \\ \hline
\hline
\end{tabular}
\end{table*}
\end{comment}

Tables~\ref{tab:reward_sr_route_type} and~\ref{tab:length_deviation_route_type} summarize the average reward, success rate, episode length, and lane deviation for the five best-performing configurations, evaluated across four route categories: Multi-Turn, One-Turn, Straight, and Full Route. In addition, Table \ref{tab:convergence} shows the convergence step and rate for different methods.

\begin{table}[ht]
\centering
\caption{Convergence step and rate for different action‐space methods}
\label{tab:convergence}
\begin{tabular}{@{}lrr@{}}
\toprule
\textbf{Method} & \textbf{Step to Target} & \textbf{Convergence Rate} \\ 
\midrule
\textbf{Rel-0.5}                     & 1,024,000  & $9.7656\times10^{-7}$ \\
\textbf{Fix-21012}           & 1,409,024  & $7.0971\times10^{-7}$ \\
\textbf{Rel-1.0}                          & 1,662,976  & $6.0133\times10^{-7}$ \\
\textbf{F-0.5}             &   1,884,160  & $5.3074\times10^{-7}$ \\
\textbf{Dyn-0.5}                  & 2,752,512  & $3.6330\times10^{-7}$ \\
\bottomrule
\end{tabular}
\end{table}
\paragraph{Multi-Turn Scenarios}
In complex multi-turn routes, the F-0.5 configuration achieved the highest reward (48458.43) and a success rate of 50\%, while the Rel-0.5 configuration matched that success rate with a slightly lower reward but lower lane deviation (0.24), indicating smoother maneuvering. The Rel-1.0 configuration failed with a success rate of 0 and much higher lane deviation (0.52), suggesting erratic behavior. While Dynamic and Fixed performed reasonably, with Fix-21012 having the lowest lane deviation. In terms of efficiency, Rel-0.5 achieves the highest value, followed by Fix-21012. This highlights that, although all methods (except Rel-1.0) share the same success rate, these two reach the target result most quickly.

\paragraph{One-Turn Scenarios}
In one-turn routes, Dyn-0.5 achieved the highest reward (53974.43), but with only 25\% success. In contrast, Rel-0.5 offered a stronger balance—achieving 70\% success and 50941.50 reward with moderate lane deviation (0.15). Rel-1.0 again underperformed in both metrics. The Full has the highest success rate among all different configs with decent deviation (0.09). In this scenario, Rel-0.5 achieves the highest efficiency (6.84), outperforming the next best F-0.5 (3.98) by a wide margin, which shows that Rel-0.5 strikes a better balance between success rate and convergence speed.

\paragraph{Straight Scenarios}  
In straight-line driving scenarios, Dyn-0.5 and Rel-0.5 configurations achieved perfect success rates (100\%) while maintaining low lane deviations (0.12 and 0.17, respectively). Rel-0.5 also achieved the highest reward (23651.53) in this category. In contrast, F-0.5 and Rel-1.0 underperformed in terms of success (50\% and 40\%, respectively), with Rel-1.0 again showing the highest lane deviation (0.40). In the straight scenario, Rel-0.5 achieves the highest efficiency (9.77), followed by Fix-21012 (5.32) and Dyn-0.5 (3.63). This demonstrates that Rel-0.5 maintains the best balance between convergence speed and success rate in this scenario as well. This result reinforces the advantage of well-structured or constrained action spaces in simple driving tasks.

\paragraph{Full Route}  
Across the full-route scenario, which includes a mix of straights and turns, the F-0.5 configuration achieved the highest reward (137167.49) and a perfect success rate (100\%). Both Dyn-0.5 and Fix-21012 configurations also reached 100\% success with slightly lower rewards and low lane deviations (0.08 and 0.11, respectively). Notably, Rel-0.5 exhibited the lowest lane deviation (0.07), but its success rate dropped to 80\%, suggesting that although it drove more precisely, it occasionally failed to complete the route. Meanwhile, Rel-1.0 had both the longest episode length (1462.60) and the highest deviation (0.17), combined with a significantly lower success rate (40\%), indicating a tendency to wander or struggle with route completion. However, in terms of efficiency, Rel-0.5 still led with a score of 7.81, outperforming all other methods despite their flawless success rates..

To summarize, episode length is best interpreted in conjunction with success rate: a long episode length with high success indicates careful, consistent driving, while long episodes with low success may point to the agent wandering aimlessly or failing to reach the goal. Similarly, low lane deviation reflects tighter adherence to the desired trajectory, which is especially important in urban driving.

\subsection{Key Insights and Takeaways}

The results reveal that no single action space configuration is optimal across all criteria; instead, each offers distinct trade-offs. Among the five tested, the Dyn-0.5 demonstrates the most consistent driving stability, exhibiting low lane deviation across most routes and competitive performance in both reward and success rate. Rel-0.5 also stands out for its balance between success rate and trajectory precision, making it a reliable choice in structured or moderately complex routes. In contrast, while the F-0.5 occasionally yields the highest reward, its performance is less stable in terms of control accuracy, especially in multi-turn scenarios. The Fix-21012 performs reliably but conservatively, often underachieving in reward, and the Rel-1.0 configuration consistently underperforms due to its instability and imprecision. Overall, dynamic and moderate relative reductions appear to provide the best compromise between learning efficiency, safe control, and route completion—particularly when robustness and stability are prioritized.

\section{Conclusion and Future Work}

In this work, we systematically investigated the impact of action space design on the performance of RL agents for AV. We introduced two novel strategies—dynamic masking and relative action space reduction—and compared them against two baselines: full action spaces and fixed discrete reductions. Across nine distinct configurations, we demonstrated that careful structuring of the action space can significantly enhance training efficiency, policy stability, and route completion rates. While full action spaces may offer higher potential rewards in certain scenarios, they often lead to increased control variance and less stable behavior. In contrast, our proposed dynamic and relative strategies strike a more favorable balance between control precision and reward acquisition, particularly in complex and structured driving environments.

For future work, we plan to extend this study in several directions. First, we aim to assess the generalization capabilities of the top-performing configurations across a broader range of scenarios in the CARLA simulator, including variations in towns, traffic densities, and weather conditions. Second, we intend to explore adaptive or learned action masking strategies, where the action space evolves dynamically based on state feedback or policy progression during training. Additionally, integrating action space reduction with safety-aware planning modules may further enhance robustness and real-world applicability.

\bibliographystyle{plain} % Options: plain, abbrv, alpha, etc.
%\bibliographystyle{IEEEtran}
%\bibliography{Action_space_reduction_for_RL,CARLA_RL_Review_Paper}
\bibliography{main}
\end{document}